# A review of TinyML

Harsha Yelchuri
Information Science Engineering
RV College of Engineering
Bengaluru, India
harshayelchuri2000@gmail.com

Rashmi R
Information Science Engineering
RV College of Engineering
Bengaluru, India
rashmir@rvce.edu.in

*Abstract*— In this current technological world, the application of machine learning is becoming ubiquitous. Incorporating machine learning algorithms on extremely low-power and inexpensive embedded devices at the edge level is now possible due to the combination of the Internet of Things (IoT) and edge computing. To estimate an outcome, traditional machine learning demands vast amounts of resources. The TinyML concept for embedded machine learning attempts to push such diversity from usual high-end approaches to low-end applications. TinyML is a rapidly expanding interdisciplinary topic at the convergence of machine learning, software, and hardware centered on deploying deep neural network models on embedded (micro-controller-driven) systems. TinyML will pave the way for novel edge-level services and applications that survive on distributed edge inferring and independent decision-making rather than server computation. In this paper, we explore TinyML's methodology, how TinyML can benefit a few specific industrial fields, its obstacles, and its future scope.

*Keywords—TinyML, machine learning*

## I. Introduction

Concurrent to the growth of the machine learning and Artificial intelligence sectors, the IoT industry increased as well, fueled by breakthroughs in interconnectivity and the availability of system-on-board solutions [1]. The percentage of microcomputer (MCU)-powered products is increasing substantially nowadays [2].

Unfortunately, in several circumstances, embedded systems will not handle the information that is captured; rather, the information is transported to a distant site for storing as well as further analysis. This might result in undesirable information delay, information leakage, and information confidentiality issues in specific applications. This sparked study on the basic difficulties in delivering machine learning techniques to embedded systems with minimal support (in terms of processing speed, storage and power consumption)

Embedded systems often obtain information from a variety of sensing devices (Infrared cameras, audio, inertial measurement units, thermocouples, gas sensors, gyroscopes, and so on). These embedded systems are frequently powered by batteries and operate at low power levels (1 mW or less). Because it often accomplishes simple operations, the device's CPU is frequently underused across time. Because embedded Microcontroller devices are widely used, their closeness to actual pieces of knowledge via sensing devices, and underused computational power, they are ideal for implementing compact inference algorithms. Tiny machine learning (TinyML) is indeed a developing embedded technology discipline that intends to apply machine learning techniques on resource-restricted, Microcontroller devices as a result of this shift.

The basic difficulties of making complex machine learning techniques and algorithms behind the severely restricted infrastructure of embedded systems and accomplishing extremely reliable data interpretation tasks are addressed by the TinyML group. With substantially less resources, TinyML seeks to identify ways to modify existing deep neural networks for usage on Microcontroller based embedded systems. It also seeks to provide supporting strategies that make model installation simple and accurate.

TinyML will support advancements in a number of industries, including electronic devices, medical services, networked cyber-physical technologies, and automation technologies. However, accomplishing these objectives necessitates interdisciplinary approaches from a variety of fields, such as hardware development, signal computation, systems design, and machine learning. In order to set the objectives and tackle the issues of this new innovative subject, the microcontrollers industry and the machine learning industry must engage with each other in the construction of TinyML. The TinyML system, which aspires to deliver Intelligence to the tiniest, least developed edge-level devices, is evaluated in this paper for its top-of-the-line.

## II. Workflow

The typical deployment process for a TinyML application may be seen in Figure 1. It is possible to partition the design into conventional ML and TinyML modules [3]. The first includes the standard steps of classic ML algorithms, such as gathering data, algorithm selection, model construction, and optimization. The model porting and deployment stages make up the latter, which focuses on TinyML.

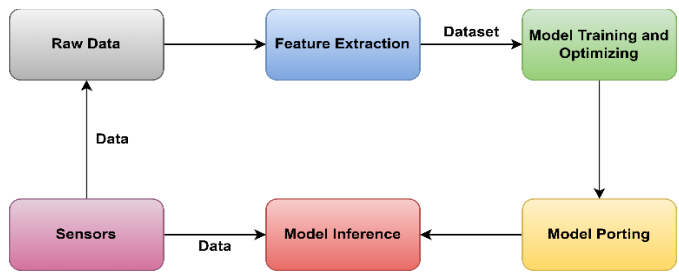

Figure 1: Generic TinyML Pipeline

The first stage in creating a TinyML model is to choose an effective algorithm and gather the necessary information. The use of pre - compiled datasets or real-time sensor device data collection can achieve this goal. After all of the requirements have been acquired, the model learning procedure may be carried out in a resource-rich device. After the model learning is done, the resulting model must be enhanced.

The last phases of a TinyML system are porting and deploying. When incorporating a model, it must be translated into an MCU-friendly language. TinyML interpreters (for example, Tensorflow Lite Micro) are utilised in this situation to transform a model developed in an appropriate scripting language to a frozen graph optimally expressed as a C array. Lastly, the ported model is incorporated in a Microcontroller, where inference on sensor information is conducted using the base ML process.

### III. TINYML USE CAESE AND APPLICATIONS

The enormous usage scope made available by the expansion of TinyML is fundamentally what propels TinyML's unparalleled rise [4]. Applications create new opportunities for even more effective TinyML frameworks [5] as well as progress across the entire development pipeline. Moreover, applications create the required data samples and the related capital growth that fuels future work. TinyML could be used to showcase a variety of applications [6]. A few examples of usage scenarios are: face identification, hand motion recognition, traffic monitoring, driverless car supervising, speech to text conversion and many others. When it comes to application fields, tinyML has now been effectively implemented along a variety of applications, including those in the areas of medicine, surveillance systems, smart things (from sensing devices to metropolitan areas), industrial predictive maintenance and control, and several other areas which are encountered in our daily societal life. The following is a tiny selection, including some prospects of TinyML in various domains [7]:

*A. Healthcare*

With the availability of intelligent economical technologies, healthcare is an industry with enormous development prospects. Wearable technology now controls the industry for low-cost health surveillance. However, present systems have limitations like the dependency on the main device (for example, a mobile phone) and the substantial power expenditure required to sense transport information, and get estimates in a wireless network, indicating why several products are rechargeable. Furthermore, the delicacy of health records necessitates procedures to assure adequate confidentiality, particularly in wireless networks, a characteristic that scarce resources may impede in wearable technology. Contrarily, TinyML is able to localize the learning and interpretation of sensor information, making way to novel solutions that may be used for longer durations without requiring regular recharging.

*B. Surveillance and security*

The main areas of concentration for TinyML at the moment are perhaps the creation of neural network models for machine vision and auditory perception, that have been the basis of surveillance. Executing TinyML programs on microcontroller devices enables operations like object recognition, face identification, basic action recognition, and speech sensing to be carried out with great precision. TinyML will make it possible to create brand-new classes of widespread, compact, and inexpensive identifying, tracking, and surveillance applications. This is especially well suited for situations in which a straightforward classification is required (such as whether a person was spotted or otherwise), rather than extensive relevant data.

*C. Augmented/Virtual Reality*

In order to create an enabling environment, virtual reality (VR) and augmented reality (AR) systems immerse users in fully or partially virtual or semi-virtual interactive spaces. AR includes three conditions that must be met: 1) The capacity to mix actual items with virtual items. 2) The capacity to synchronize actual and digital items. 3) Real-time interaction with the augmented reality environment through motion detection etc. Modern gesture-based interacting AR/VR devices are frequently an upscale indulgence that prevents the mass adoption of such technology. Considering these, the study tries to significantly decrease the price of these technologies by proposing inexpensive equipment replacements for costly equipment that has adequate computing power. Additionally, TinyML has an edge over competing studies since it does not require the computing power of a machine with rich resources.

*D. Industrial maintenance*

As a result of improvements in the automated systems of typical manufacturing divisions, improved networking, and intelligent sensing technologies, Industrial IoT practices expand quickly. Further improvements to these procedures can be supported by TinyML capabilities. Predictive data maintenance may be used on several industrial equipments on embedded hardware. TinyML can help several industrial monitoring systems that have time-sensitive requirements.

### IV. CHALLENGES

TinyML is still in its growing stage, and this area of study and innovation is expanding quickly. The advancements in the nexus of several disciplines, particularly ML algorithms, development environments, and hardware compatibility, will guarantee the effectiveness of this area of research, which is being conducted in many different ways. This technology encounters a variety of obstacles along the road, like as [8][9]:

*A. Low power*

TinyML's main area of study concern is the little power that is available on edge devices. The energy-saving functionality must be supported by TinyML's philosophy. TinyML models need a specific amount of power to keep the model's efficiency at a given level. Given the diversity in their architecture and pre-processing routes, hardware modules make it challenging to define a standard power mechanism. As sensors and other accessories are typically attached to edge devices, TinyML frameworks may experience some energy mishandling. Consequently, creating a power-effecient TinyML system continues to be a major obstacle.

*B. Limited memory*

This is another barrier preventing TinyML from expanding. Another vertical of resistance is what prevents TinyML from expanding. It is particularly challenging to implement ml algorithms due to the extremely low size of SRAM. For example, GB-sized storage peaks may be used for traditional ml inference, which is not feasible for edge systems. As a result of edge equipment's limited memory, the development of TinyML frameworks becomes difficult.

*C. Hardware and Software Heterogeneity*

TinyML technologies have a wide range of performance, energy, and capacities considering their infancy. The devices span from standard MCUs to cutting-edge designs. Due to the fact that the device undergoing test (SUT) may not always have characteristics that are generally considered standard, such as a clock, this diversity presents a myriad of issues. Another difficult problem is standardizing performance findings across various configurations.

Model installation on TinyML systems may be done in three different ways: manually, code generation, or via ML interpreters. The optimum outcomes are frequently achieved by manual writing since it enables minimal, implementation-specific optimization; nevertheless, the process is extremely time-consuming and the effects of the performance improvements are frequently ambiguous. Code generation techniques generate well-optimized software without the labor-intensive hand-coding process. Nevertheless, because each major manufacturer has their own set of exclusive methods and compilers. Code generation would not solve the difficulties with comparison. This also causes porting difficult. Since ML interpreters' abstract patterns are consistent throughout platforms, they offer great portability. The speed and binary size expenses for this solution are moderate.

*D. Absence of benchmarking tools*

To fully comprehend TinyML efficiency, microcontrollers executing TinyML-based machine learning techniques have to be evaluated. To assess, examine, and consistently record distinct performance variances across systems, a collection of benchmarking resources and techniques are needed.

## V. FUTURE ROAD MAP

The plethora of difficulties mentioned previously, gives a large number of chances for exploration and development. Solving these difficulties demands interdisciplinary approaches that span on all layers, from programming techniques to micro-architecture to hardware [10]. A few of the sectors with potential for growth in the future are:

*A. Task Offloading*

For edge-enabled computational contexts, the on/off-loading of computing tasks may be utilised. Loading tactics of this type must be added to the current flexible setup of edge-aware details. TinyML could thus enable the movement of resource-intensive operations from the resource-constrained edge of the network. Connectivity between both the edge and the cloud may therefore be highly utilized. An examination should be conducted to discover the core procedures underpinning these resource utilization.

*B. Limited memory Communication techniques*

MCUs can run unassisted utilising power supply or ambient energy collection techniques. Recurring actions like model exchanges are the foundation of the durability assurances for many of these devices in terms of network-dependent reformability. The amount of information which ought to be exchanged varies depending on the size of the model. As a result, it influences the amount of time and energy required for communication activities. Considering the potential network scenarios caused by the positioning of MCUs, minimising unnecessary interactions and data retransmission is crucial. For this reason, we propose that concise intelligence techniques on the Higher layers be developed to utilise modern RATs like BLE or LoRa.

*C. Modern computing concepts*

In-memory processing, for instance, has produced good outcomes for machine learning problems, but it still needs more study, comprehension, and improvement before TinyML systems can use it. The reconfigurable fabric has also demonstrated considerable promise as a specialised hardware accelerators for ML inferences, allowing customers to optimise hardware for each purpose. Reconfigurable fabric is not a choice right now, given the size of tinyML-targeted devices and the needs for its energy usage, related programming equipment, I/O, and some other factors.

## CONCLUSION

The potential to use intelligent strategic decisions at the edge level is enormous for the edge-IoT platform. Incorporating ml algorithms in compact embedded devices with limited resources is becoming increasingly necessary in light of upcoming application domains. In-depth research into the TinyML conceptual framework is required to improve the present edge-aware device learning environment. This paper examines the fundamentals of TinyML. In addition, the various TinyML use cases, application domains, and difficulties in creating TinyML applications are examined. The TinyML roadmap for the future is also explored.